\documentclass[11pt,a4paper]{article}
\usepackage[utf8]{inputenc}
\usepackage[T1]{fontenc}
\usepackage[english]{babel} 
\usepackage{amsmath,amssymb,amsfonts}
\usepackage{mathtools}
\usepackage{graphicx}
\usepackage{hyperref}
\usepackage{xcolor}
\usepackage{booktabs}
\usepackage{multirow}
\usepackage{microtype}
\usepackage{setspace}
\usepackage{placeins}
\usepackage[a4paper,margin=2.2cm]{geometry}
\usepackage{tikz}
\usetikzlibrary{arrows.meta,positioning,calc,fit}
\usepackage{csquotes}

\hypersetup{colorlinks=true,linkcolor=blue!70!black,citecolor=blue!70!black,urlcolor=blue!70!black}

\title{Differentiable Tripartite Modularity for Clustering Heterogeneous Graphs}
\author{Beno\^it Hurpeau}

\begin{document}
\maketitle
\begin{center}
\end{center}

\begin{abstract}
Clustering heterogeneous relational data remains a central challenge in graph learning, particularly when interactions involve more than two types of entities. While differentiable modularity objectives such as DMoN have enabled end-to-end community detection on homogeneous and bipartite graphs, extending these approaches to higher-order relational structures remains non-trivial.

In this work, we introduce a differentiable formulation of tripartite modularity for graphs composed of three node types connected through mediated interactions. Community structure is defined in terms of weighted co-paths across the tripartite graph, together with an exact factorized computation that avoids the explicit construction of dense third-order tensors. A structural normalization at pivot nodes is introduced to control extreme degree heterogeneity and ensure stable optimization.

The resulting objective can be optimized jointly with a graph neural network in an end-to-end manner, while retaining linear complexity in the number of edges. We validate the proposed framework on large-scale urban cadastral data, where it exhibits robust convergence behavior and produces spatially coherent partitions. These results highlight differentiable tripartite modularity as a generic methodological building block for unsupervised clustering of heterogeneous graphs.
\end{abstract}
\newpage

\section{Introduction}

Modularity-based objectives have long provided an interpretable criterion for
community detection, by comparing observed connectivity patterns to a suitable
null model \cite{newman2004,fortunato2016}. Recent advances have introduced
differentiable relaxations of modularity, such as DMoN, enabling end-to-end
optimization jointly with graph neural network encoders \cite{tsitsulin2020}.
These approaches have proven effective for homogeneous graphs and, in some
cases, bipartite structures \cite{barber2007, murata2009bip}.

However, extending modularity-based objectives to higher-order relational
structures remains challenging. Several extensions have been proposed for
multipartite or multi-relational networks, but they typically rely on discrete
optimization or strong structural assumptions \cite{neubauer2009}. In
particular, tripartite graphs cannot be reduced to pairwise interactions
without losing structural information, while explicit hypergraph formulations
often lead to prohibitive computational costs or non-differentiable objectives
\cite{zhou2007hypergraph, bretto2013hypergraph}.

In the field of graph learning, most deep clustering approaches for spatial or
geographical data define neighborhoods using Euclidean distances or
$k$-nearest-neighbor graphs \cite{liu2014}. Such heuristics ignore the physical
constraints that structure real-world urban spaces and may introduce artificial
connections between entities that are spatially close but topologically
disconnected.

In this work, we address these limitations by proposing a differentiable
formulation of tripartite modularity tailored to graphs composed of three node
types connected through mediated interactions. Our approach defines community
structure in terms of weighted co-paths across the tripartite graph and relies
on a factorized computation that avoids the explicit construction of dense
third-order tensors. A structural normalization at the pivot nodes is introduced
to handle extreme degree heterogeneity and ensure stable optimization, in line
with previous extensions of modularity to heterogeneous settings
\cite{murata2011}.

\section{Data and Methodological Construction}

This section describes the construction of the tripartite graph and the associated methodological choices. The graphs considered in this work fall within the broader category of heterogeneous graphs, for which numerous graph neural network architectures have been proposed in recent years \cite{zhang2019heterogeneous}. The urban data mobilized here provide a realistic validation setting, enabling the evaluation of the model’s behavior on a large heterogeneous graph, without constituting a thematic or application-driven analysis objective in itself.

\subsection{Data Sources and Graph Ontology}

The graph is constructed by integrating several French administrative registries: the cadastral register (parcels), the National Address Base (BAN), and the National Building Repository (RNB), which serves as an interoperability pivot between the different sources. These datasets are selected for their exhaustive coverage, stable structuring, and topological consistency, which are essential properties for evaluating a large-scale graph learning framework.

The resulting graph $G = (V, E)$ is composed of three distinct types of nodes:
\begin{itemize}
  \item $X$: addresses,
  \item $Y$: buildings,
  \item $Z$: cadastral parcels.
\end{itemize}

Edges are strictly typed:
\begin{itemize}
  \item $E_{XY}$: access or association relations between an address and a
  building,
  \item $E_{YZ}$: implantation relations between a building and one or more
  parcels.
\end{itemize}

No direct edge is defined between node sets $X$ and $Z$. Tripartite interactions emerge exclusively through the pivot set $Y$, in accordance with the observed structure of the data. The proposed framework therefore relies on the existence of co-paths connecting $X \rightarrow Y \rightarrow Z$. As a consequence, nodes in $X$ or $Z$ that are not connected to any pivot node in $Y$ do not participate in the tripartite dynamics and are excluded from the effective graph used for learning. This structural hypothesis is consistent with the modeling objective, which explicitly targets interactions mediated by pivot nodes.

\subsection{Limitations of Geometric Neighborhoods}

In many works in geospatial graph learning, the neighborhood is defined based on Euclidean proximity criteria, typically via $k$-nearest neighbor ($k$-NN) graphs.  While these constructions are simple to implement, they present a major limitation in an urban context: they ignore the real physical constraints that structure the space.  Two spatially close entities may be separated by a strong topological discontinuity (roadway, distinct land ownership), while more distant entities may be functionally and morphologically connected.  The use of purely geometric neighborhoods thus introduces artificial edges that do not correspond to observable physical relations, biasing the input graph structure. 

\subsection{Topological Construction Constrained by the Cadastre}

To avoid these biases, the graph topology is derived directly from the cadastral structure. Relations between parcels are defined based on their physical contiguity, weighted by the length of the shared boundary. For two parcels $P_i$ and $P_j$, the edge weight is defined as:
\begin{equation}\label{eq:wZZ}
w^{ZZ}_{ij} = \log\left(1 + \text{length}(\partial P_i \cap \partial P_j)\right).
\end{equation}

The transformation $\log(1+\ell)$ applied to lengths aims to compress the weight dynamics while preserving their relative ordering. It prevents a small number of exceptionally long edges from dominating the aggregation process, without introducing an arbitrary threshold. Other monotonic increasing functions with a similar saturation effect lead to comparable behaviors; this choice is not critical for the proposed method.

Connectivity between buildings is then obtained through a dual topological projection of the parcel graph. Let $M^{YZ}$ denote the building--parcel incidence matrix; the resulting topological adjacency between buildings is given by:
\begin{equation}\label{eq:AYY}
A^{YY}_{\text{topo}} = M^{YZ} \cdot A^{ZZ} \cdot (M^{YZ})^{\top}.
\end{equation}

\begin{figure}[htbp]
\centering
    \includegraphics[width=0.75\linewidth]{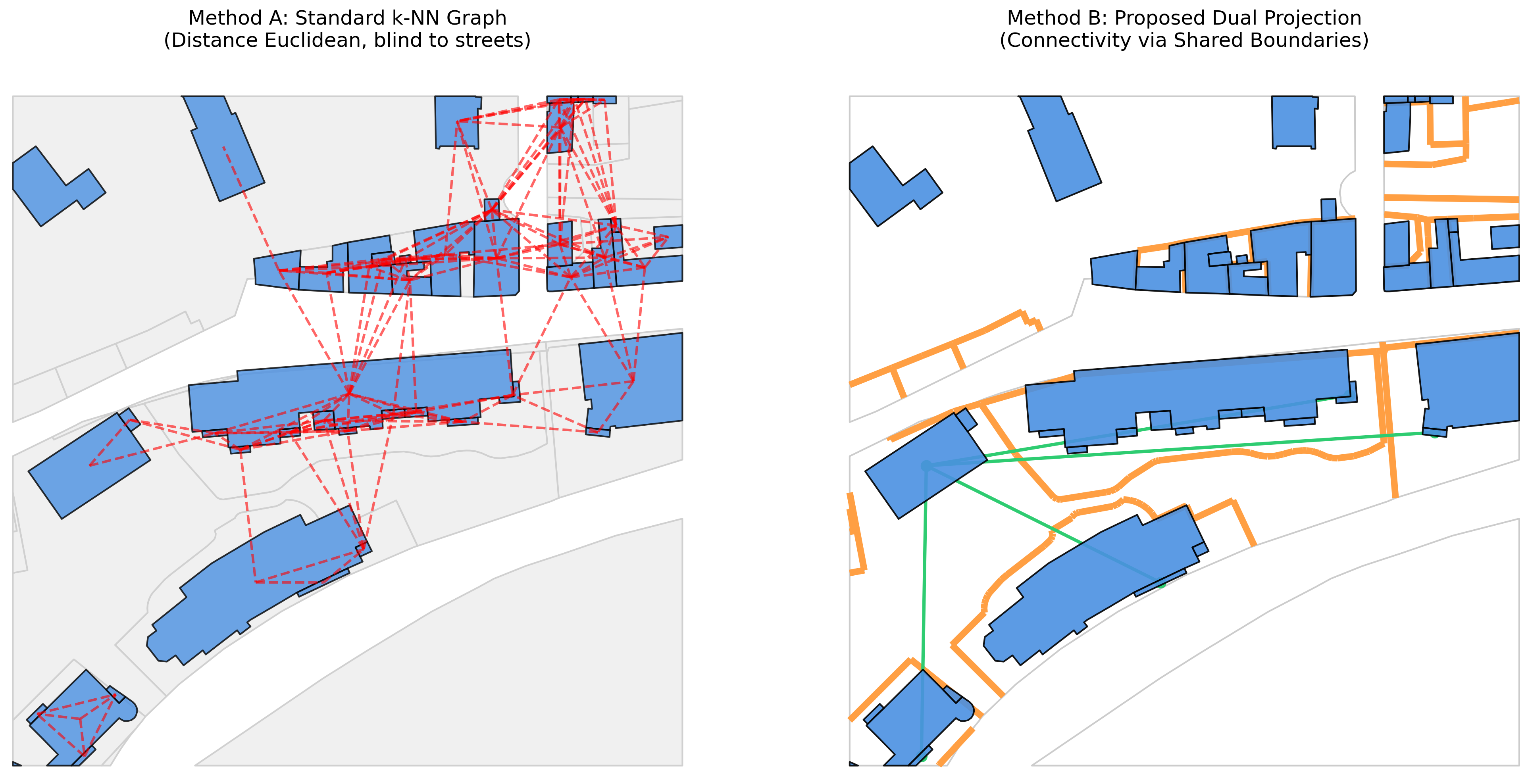}
    \caption{Comparison of graph construction strategies on a residential area.  Left (Method A): Standard k-NN graph (k=8) based on Euclidean distance.  Note that edges (red dotted lines) arbitrarily cross the roadway, creating artificial connections between distinct urban blocks.  Right (Method B): The proposed Dual Topological Projection.  Orange lines represent common parcel boundaries (weighted by length, Eq.~\ref{eq:wZZ}).  Green lines represent the resulting connectivity between buildings. The topology strictly respects the street layout: no edge connects separated blocks, despite their geographic proximity.}
    \label{fig:gnn}
\end{figure}

This construction guarantees that two buildings are connected only if they rest on parcels that are themselves connected (figure~\ref{fig:gnn}), thereby preventing the creation of edges that arbitrarily cross public space. The projection is implemented using sparse matrix operations, ensuring linear complexity in the number of edges.

\subsection{Attribute and Position Encoding}

Each building-type node is associated with an attribute vector describing elementary geometric properties (area, compactness, elongation, convexity).  These descriptors are chosen for their scale invariance and their ability to differentiate built forms without resorting to explicit functional categories.  To introduce absolute position information while preserving the graph structure, building centroids are projected via a Fourier positional encoding, following a standard formulation:
\begin{equation}\label{eq:fourier}
\gamma(p) = [\sin(2\pi \omega_1 p), \cos(2\pi \omega_1 p), \ldots,
\sin(2\pi \omega_k p), \cos(2\pi \omega_k p)],
\end{equation}
where $p = (x,y)$ denotes the node position and $\{\omega_k\}$ a set of fixed frequencies. 

\subsection{Data Status}

The urban data mobilized in this study are used exclusively as an experimental support allowing evaluation of:
\begin{itemize}
  \item the numerical stability of the model,
  \item its capacity to avoid partition collapse,
  \item its scalability on large heterogeneous graphs. 
\end{itemize}

No thematic or typological interpretation of the partitions is sought at this stage.  The choices of data and topological construction aim solely to provide a realistic and constraining use case for the evaluation of the proposed methodological framework. 
\FloatBarrier

\section{Differentiable Tripartite Modularity}

This section introduces the theoretical formulation of differentiable tripartite modularity and its integration into an end-to-end learning framework. The objective is to define a modularity-based loss function that can be optimized directly on heterogeneous tripartite graphs, without resorting to bipartite projections or discrete community assignments.

\subsection{Background on Modularity for Heterogeneous Graphs}

Modularity is a classical measure for assessing the quality of a graph partition, based on the comparison between observed intra-community connectivity and its expectation under a suitable null model \cite{newman2004,fortunato2016}. Extensions to bipartite graphs have been proposed by Barber \cite{barber2007} and Murata \cite{murata2009bip}, relying on incidence matrices and null models adapted to multi-set structures.

More generally, several works have emphasized the importance of higher-order connectivity patterns in networks, beyond pairwise interactions \cite{benson2016higher}. However, such approaches typically rely on discrete motif counts or combinatorial constructions, which remain difficult to integrate into differentiable learning frameworks.

In the tripartite setting, a direct formulation of modularity requires the joint consideration of interactions across three distinct node sets. Existing formulations remain largely non-differentiable and poorly suited for end-to-end optimization.

\subsection{Tripartite Modularity via Co-paths}

Let $G = (X \cup Y \cup Z, E_{XY} \cup E_{YZ})$ be a tripartite graph in which interactions between node sets $X$ and $Z$ are exclusively mediated by the pivot set $Y$. The tripartite structure can thus be represented through co-paths of the form $X \rightarrow Y \rightarrow Z$, obtained by composing the corresponding incidence relations.

Alternative formulations based on explicit hypergraph representations or higher-order neural architectures have been proposed to model such interactions \cite{feng2019hypergraph}. While expressive, these approaches typically incur substantial computational overhead and rely on dense higher-order tensors.

In contrast, we define tripartite modularity directly in terms of mediated co-paths, preserving the underlying relational semantics while enabling a factorized and scalable computation.

\subsection{Co-path Tensor and Density Correction}

The major challenge of geospatial graphs lies in their extreme degree heterogeneity, often following a power-law distribution: a vertical condominium may connect hundreds of addresses to a single parcel. Without explicit correction, such highly connected pivot nodes artificially induce large modular contributions solely through edge volume, biasing modularity maximization independently of any meaningful relational structure.

To neutralize this bias, we define the tripartite adjacency tensor $A$ not as a simple binary product, but as a normalized co-path density:

\begin{equation} \label{eq:A-norm-rework}
A(i,j,k) = \underbrace{\mathbb{I}\{(i,j)\in E^{XY}\}\cdot \mathbb{I}\{(j,k)\in E^{YZ}\}}_{\text{Path existence}} \times \underbrace{\frac{w^{XY}_{ij}\,w^{YZ}_{jk}}{\deg_X(j)\,\deg_Z(j)}}_{\text{Local flow normalization}},
\end{equation}
where $\deg_X(j)$ and $\deg_Z(j)$ are the weighted degrees of building $j$ towards $X$ and $Z$.  The weighting $\omega_j = (\deg_X(j)\deg_Z(j))^{-1}$ acts as a structural normalization mechanism limiting the disproportionate influence of high-degree nodes on the optimization objective.  The terms $\deg_X(j)$ and $\deg_Z(j)$ denote the weighted degrees of the pivot node $j$ towards sets $X$ and $Z$, defined respectively by:
\begin{equation}
\deg_X(j) = \sum_{i \sim j} w^{XY}_{ij}, \qquad
\deg_Z(j) = \sum_{k \sim j} w^{YZ}_{jk}. 
\end{equation}
These quantities correspond to the total mass of incoming and outgoing flux of pivot $j$ in the associated bipartite graph.  Expression~\eqref{eq:A-norm-rework} formally defines the weight associated with a co-path $(i,j,k)$, but does not correspond to a dense tensor explicitly instantiated in memory.  In practice, the tripartite tensor $A$ is never constructed: only the factorization by pivot described in Section 3.7 is implemented, allowing exact calculation of triadic fractions with linear complexity in the number of edges (\ref{sec:complexity}). 

\subsection{Differentiable Relaxation and Factorization}

To make this modularity optimizable via gradient descent, we introduce a continuous relaxation of the assignment matrices.  Each node is associated with a probabilistic membership vector to a set of communities, of sizes $(L,M,N)$ depending on its type $(X,Y,Z)$, grouped in matrices $S_X$, $S_Y$, and $S_Z$.  The tripartite contribution can then be rewritten in factorized form, avoiding the explicit manipulation of a rank-three tensor.  This factorization relies on an exact rewriting of co-path terms using successive matrix products, allowing efficient implementation using sparse operations.  This step constitutes a key point of the proposed framework: it enables end-to-end optimization of tripartite modularity while maintaining linear complexity in the number of graph edges. 

\subsection{Flow Normalization}

Tripartite urban graphs are characterized by extreme degree heterogeneity, particularly at the level of pivot nodes.  Without explicit correction, modularity maximization leads to partition collapse, dominated by a small number of very high-degree nodes.  To mitigate this phenomenon, we introduce a flow normalization weighting the contribution of pivot nodes according to their degree.  This weighting aligns with generalizations of modularity to heterogeneous and tripartite graphs \cite{murata2011} and acts as a normalization term limiting the disproportionate influence of hubs.  The introduction of this term is indispensable for the numerical stability of the model, as shown by the ablation study presented in Section~\ref{sec:experiments}. 

\subsection{Link to DMoN}

The proposed formulation aligns with works on differentiable modularity introduced by DMoN \cite{tsitsulin2020}.  Unlike DMoN, initially designed for homogeneous or bipartite graphs, the framework presented here operates directly on tripartite graphs, without prior reduction of their structure.  Differentiable tripartite modularity can thus be optimized jointly with a graph encoder, in an end-to-end framework, while retaining an explicit interpretation of the objective function. 

\subsection{Factorized Aggregation of Triadic Fractions}

The observed triadic fraction $e_{lmn}$ measures the relative density of co-paths connecting address community $l$, building community $m$, and parcel community $n$. A naïve evaluation based on the explicit tripartite adjacency tensor $A$ would incur a prohibitive computational complexity of $\mathcal{O}(|X|\cdot|Y|\cdot|Z|)$.

However, the local structure induced by the normalized co-path definition in Eq.~\eqref{eq:A-norm-rework} enables an exact factorized reformulation centered on the pivot nodes $j \in V^Y$. Using the assignment matrices $S_T$ produced by the GNN, the triadic fraction can be written as:
\begin{equation}
e_{lmn} = \frac{1}{M} \sum_{j \in V^Y} \omega_j \cdot \Psi_{j}(l, m, n),
\label{eq:elmn_factor}
\end{equation}
where $\Psi_j$ aggregates the contributions of the upstream and downstream
neighbors of pivot $j$:
\begin{equation}
\Psi_{j}(l, m, n) =
\underbrace{\left(\sum_{i \sim j} w^{XY}_{ij} S_X[i,l]\right)}_{A_{j,l}}
\cdot S_Y[j,m] \cdot
\underbrace{\left(\sum_{k \sim j} w^{YZ}_{jk} S_Z[k,n]\right)}_{C_{j,n}}.
\end{equation}

The terms $A_{j,l}$ and $C_{j,n}$ are precomputed via simple message-passing operations (scatter-add type). This formulation reduces the overall computational complexity to $\mathcal{O}(|E^{XY}| + |E^{YZ}|)$, making the approach compatible with large-scale heterogeneous urban graphs.

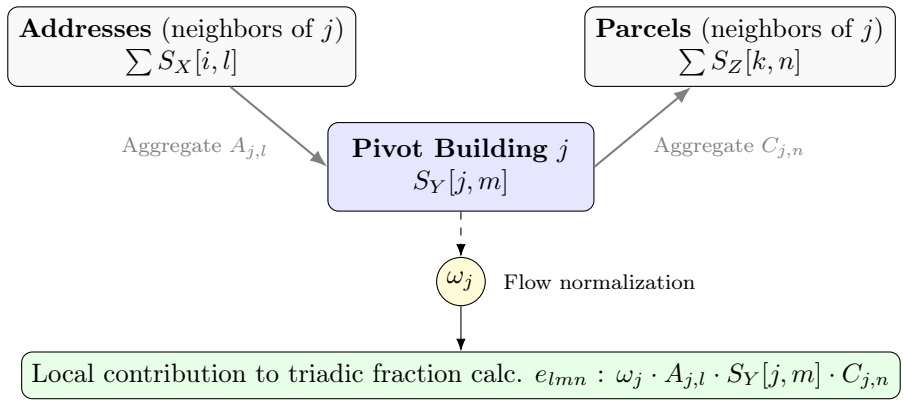
\begin{figure}[htbp]
\centering
\begin{tikzpicture}[
  >=Latex, node distance=1.2cm, every node/.style={font=\small},
  sbox/.style={draw,rounded corners,fill=gray!5,align=center,inner sep=4pt,minimum width=32mm},
  eqbox/.style={draw=none,fill=white,align=left,inner sep=2pt,font=\scriptsize,text width=45mm}
]
\node[sbox] (Xblk) {\textbf{Addresses} (neighbors of $j$)\\$\sum S_X[i,l]$};
\node[sbox, right=30mm of Xblk] (Zblk) {\textbf{Parcels} (neighbors of $j$)\\$\sum S_Z[k,n]$};
\path let \p1 = (Xblk), \p2 = (Zblk) in coordinate (midXZ) at ($(\p1)!0.5!(\p2)$);
\node[draw,rounded corners,fill=blue!10,align=center,inner sep=6pt, minimum width=35mm, below=1.0cm of midXZ] (Bj) 
    {\textbf{Pivot Building} $j$\\ $S_Y[j,m]$};
\draw[-{Latex}, thick, gray] (Xblk) -- (Bj.west) node[midway, below left, font=\scriptsize] {Aggregate $A_{j,l}$};
\draw[{Latex}-, thick, gray] (Zblk) -- (Bj.east) node[midway, below right, font=\scriptsize] {Aggregate $C_{j,n}$};
\node[draw, circle, fill=yellow!20, inner sep=2pt, below=0.6cm of Bj] (omega) {$\omega_j$};
\draw[->, dashed] (Bj) -- (omega);
\node[right=0.1cm of omega, font=\scriptsize, align=left] {Flow normalization};

\node[draw, rounded corners, fill=green!10, below=0.6cm of omega, minimum width=6cm] (res) 
    {Local contribution to triadic fraction calc. $e_{lmn}$ : $\omega_j \cdot A_{j,l} \cdot S_Y[j,m] \cdot C_{j,n}$};
\draw[->] (omega) -- (res);

\end{tikzpicture}
\caption{Diagram of the factorized modularity calculation.  The pivot $j$ aggregates the assignments of its neighbors ($A_{j,l}, C_{j,n}$), weights them by its own assignment ($S_Y$) and its correction factor ($\omega_j$).}
\label{fig:factorisation}
\end{figure}

Figure~\ref{fig:factorisation} illustrates this factorized computation: for each pivot node $j$, upstream and downstream contributions are aggregated independently, weighted by the pivot’s assignment and its structural normalization factor, and then combined to produce the local contribution to the triadic fraction $e_{lmn}$.

\FloatBarrier

\section{End-to-End Optimization: DMoN-3p}

\subsection{From Murata to a \emph{Soft} Formulation}
Classical tripartite modularity in its discrete formulation (Murata \cite{murata2011}) relies on a hard correspondence between communities: each community $l$ of $X$ is paired with a unique community $m^*(l)$ of $Y$. This operation, implemented via an $\arg\max$, is non-differentiable and prevents gradient backpropagation to the GNN.

Related differentiable relaxation approaches have been proposed for graph clustering, notably through learned hierarchical pooling mechanisms \cite{ying2018hierarchical}. However, these methods are formulated for homogeneous graphs and do not explicitly address heterogeneous tripartite structures.

To enable end-to-end learning, we replace hard matching with a probabilistic correspondence (\emph{soft matching}). We first compute the bipartite confusion matrices $E^{XY}_{lm} = \sum_n e_{lmn}$ and $E^{YZ}_{mn} = \sum_l e_{lmn}$. The matching weights are then obtained via a softmax operation controlled by a temperature parameter $\beta$:
\begin{equation}
\alpha_{m,l} = \frac{\exp(\beta E^{XY}_{lm})}{\sum_{l'} \exp(\beta E^{XY}_{l'm})},
\qquad
\gamma_{m,n} = \frac{\exp(\beta E^{YZ}_{mn})}{\sum_{n'} \exp(\beta E^{YZ}_{mn'})}.
\end{equation}
Here, $\alpha_{m,l}$ (resp. $\gamma_{m,n}$) encodes the probability of correspondence between community $l$ of $X$ and community $m$ of $Y$ (resp. between $m$ and $n$).

\noindent
The parameter $\beta$ plays a role analogous to an inverse temperature:
\begin{itemize}
    \item $\beta \to \infty$: convergence toward Murata’s hard assignment,
    recovering a strict one-to-one correspondence;
    \item $\beta \to 0$: uniform mixing, corresponding to maximum entropy.
\end{itemize}
In practice, a moderate value of $\beta$ allows gradients to propagate across multiple candidate correspondences during the early stages of training, thereby stabilizing convergence.

\subsection{DMoN-3p Loss Function}

The loss function to be minimized combines the \emph{soft} tripartite modularity
$Q_{\text{tri-soft}}$ with a collapse regularization term
$\mathcal{R}_{\text{col}}$ applied to each node type:
\begin{equation}
\mathcal{L}_{\text{DMoN-3p}} = - Q_{\text{tri-soft}} +
\sum_{T \in \{X,Y,Z\}} \lambda_T \mathcal{R}_{\text{col}}(S_T).
\end{equation}

The soft tripartite modularity is defined as:
\begin{equation}
Q_{\text{tri-soft}} =
\sum_{l,m,n} (\alpha_{m,l}\gamma_{m,n})
\left( e_{lmn} -
\underbrace{a^X_l a^Y_m a^Z_n}_{\text{Null model}} \right).
\end{equation}

The terms $a^X_l$, $a^Y_m$, and $a^Z_n$ correspond to the marginal masses induced
by the observed triadic fraction $e_{lmn}$, defined as:
\begin{equation}
a^X_l = \sum_{m,n} e_{lmn}, \qquad
a^Y_m = \sum_{l,n} e_{lmn}, \qquad
a^Z_n = \sum_{l,m} e_{lmn}.
\end{equation}
By construction, these quantities are normalized and satisfy
$\sum_l a^X_l = \sum_m a^Y_m = \sum_n a^Z_n = 1$.

The tripartite null model is thus defined as the expected triadic fraction under the hypothesis of independent community assignments, given by the factorized form:
\begin{equation}
e^{\text{null}}_{lmn} = a^X_l \, a^Y_m \, a^Z_n.
\end{equation}

The term $\mathcal{R}_{\text{col}}$ corresponds to the collapse regularization introduced in DMoN \cite{tsitsulin2020} and is not detailed here, as its role and formulation remain unchanged. Importantly, collapse regularization plays a role that is distinct and orthogonal to the structural normalization introduced in the definition of the tripartite adjacency tensor. While structural normalization acts directly on co-path contributions by bounding their influence independently of node degree, collapse regularization intervenes exclusively at the optimization level, preventing trivial convergence of the assignment matrices toward uniform or degenerate distributions. These two mechanisms therefore address complementary issues: structural stability on the one hand, and learning stability on the other.

\section{Algorithmic Analysis and Complexity}

\subsection{Computational Complexity}\label{sec:complexity}
The main strength of the proposed framework lies in its computational efficiency.  Unlike spectral approaches on hypergraphs, which require the manipulation of dense matrices of size $N \times N$, the calculation of the objective function is here dominated by local aggregation operations and contractions over communities. 
\begin{itemize}
    \item \textbf{Aggregation ($A_{j,l}, C_{j,n}$)}: $\mathcal{O}(|E^{XY}| \cdot L + |E^{YZ}| \cdot N)$. Linear in the number of edges. 
    \item \textbf{Contraction ($e_{lmn}$)}: $\mathcal{O}(|Y| \cdot L \cdot N)$. Linear in the number of pivots. 
    \item \textbf{Modularity}: $\mathcal{O}(L \cdot M \cdot N)$. Independent of the number of graph nodes, depends only on the number of communities. 
\end{itemize}
In practice, the numbers of communities $(L,M,N)$ remain low (generally fewer than 100), which makes the approach compatible with the processing of large-scale heterogeneous graphs on standard GPU architectures. 

\subsection{Optimization Properties}

The proposed framework jointly optimizes:
\begin{enumerate}
    \item continuous latent representations of nodes, learned by the GNN from local attributes and graph structure; 
    \item probabilistic assignment matrices defining a soft partition of the three sets of nodes. 
\end{enumerate}

The weighting $\omega_j = (\deg_X(j)\deg_Z(j))^{-1}$ acts as a mechanism for flow normalization per pivot, ensuring that the total contribution of a pivot node is bounded independently of its degree.  This property contributes to the numerical stability of the optimization and the prevention of partition collapses, independently of any thematic interpretation of the learned communities. 

\section{Empirical Validation and Ablation Study}
\subsection{Experimental Protocol}

The DMoN-3p model is evaluated on the complete graph of the Hauts-de-Seine department (France, D092), constructed according to the topological methodology presented in Section II and comprising $N = 308,241$ buildings.  In the experiments, capacities $(L,M,N)$ are initialized to a maximum value $K_{\max}=64$ for each node type.  The DMoN framework does not require all these communities to be activated: during learning, the model is free to deactivate a portion of them.  A decrease in the number of communities effectively used reflects a non-trivial structuring of assignments, while maintaining uniform occupation close to $K_{\max}$ corresponds to a \emph{soft collapse} regime, in which nodes are distributed almost randomly among clusters.  Hyperparameters are optimized by Bayesian search (TPE) to maximize modularity while avoiding trivial collapse regimes.  The final model uses a learning rate $\eta \approx 4.4 \times 10^{-3}$ and a collapse penalty $\lambda_{\text{collapse}} \approx 0.068$.  To foster progressive community specialization, the inverse temperature $\beta$ follows a linear annealing from $2.0$ to $7.16$ over $150$ epochs.  The inverse temperature parameter $\beta$ controls the degree of \emph{hardening} of the probabilistic correspondence between communities.  Low values favor diffuse correspondence at the start of learning, facilitating gradient flow, while high values progressively approach the discrete behavior of Murata's original formulation.  The final value $\beta \approx 7.16$ results from Bayesian hyperparameter optimization and possesses no intrinsic interpretation;  values of the same order of magnitude lead to similar dynamics. 

\subsection{Ablation Study: Impact of Flow Normalization}
\label{sec:experiments}

To evaluate the role of pivot flow normalization, denoted by the degree weighting $\omega_j$ (Eq.~\ref{eq:A-norm-rework}), we compare the full model (DMoN-3p) to an ablated variant in which $\omega_j = 1$ for all $j \in V^Y$.  The two configurations exhibit radically different learning dynamics:

\begin{enumerate}
    \item \textbf{With Flow Normalization.}  
    The model converges to a stable partition composed of $48$ active communities.  The distribution of cluster sizes is strongly unbalanced, with one majority cluster and several secondary communities of significant sizes.  The model's logic gates remain active (mean value $\approx 0.90$), indicating effective use of representation capacity. 
    \item \textbf{Without Correction (Ablation).}  
    The model undergoes a \emph{soft collapse} regime: the average assignment of clusters becomes quasi-uniform ($\approx 1/K$), preventing any stable specialization.  This phenomenon leads to an artificial fragmentation of the partition, a classic symptom of disproportionate influence induced by extreme degree heterogeneity. 
\end{enumerate}

We define \emph{soft collapse} here as a learning dynamic in which assignment matrices converge towards quasi-uniform distributions over communities, preventing effective specialization despite a high number of active clusters.  These results confirm that the $\omega_j$ correction is indispensable for stabilizing modularity optimization in tripartite graphs with strong degree heterogeneity. 

\subsection{Partition Visualization}

\begin{figure}[tbp]
\centering
    \includegraphics[width=0.55\linewidth]{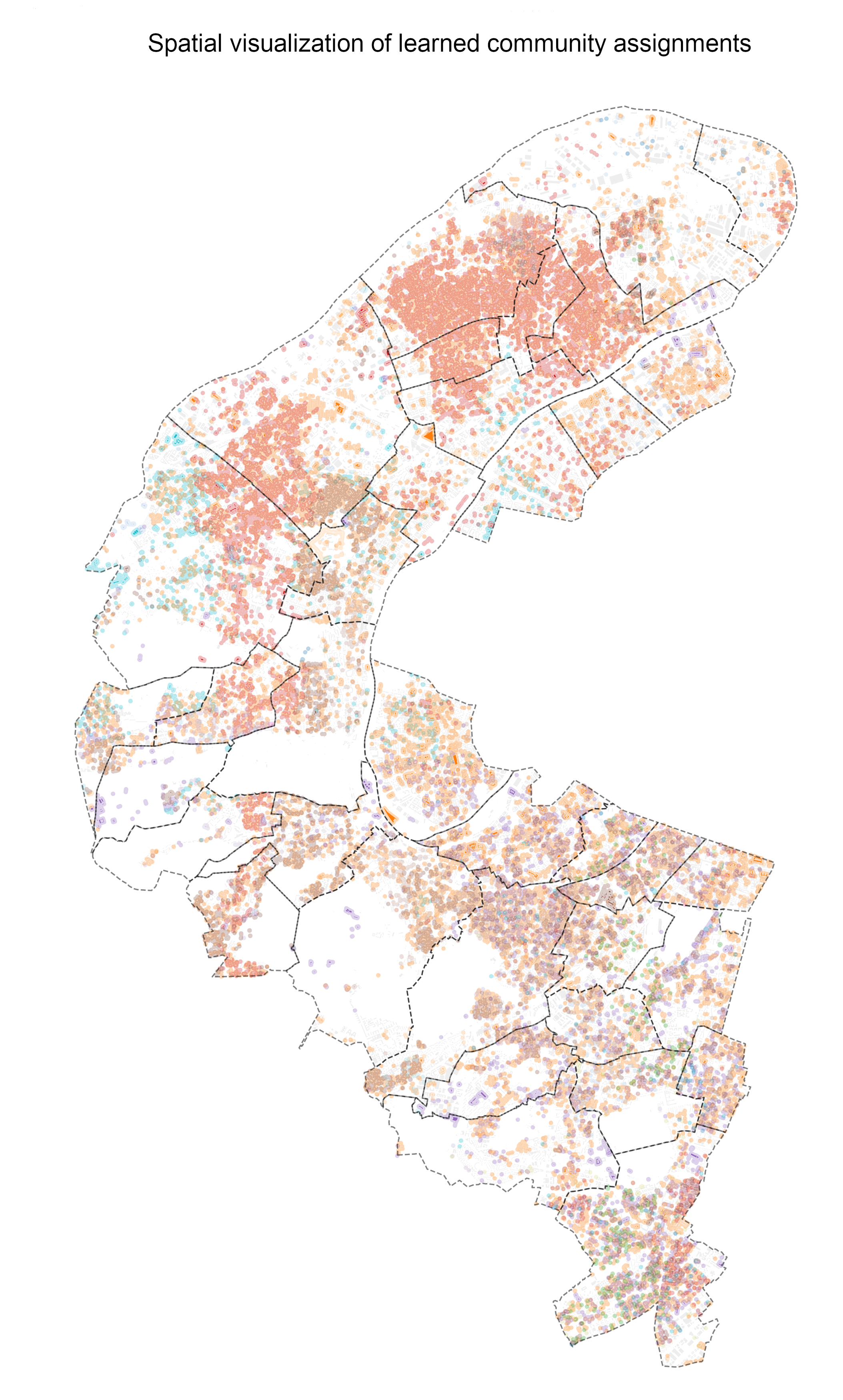}
    \caption{Illustrative mapping of communities detected by the DMoN-3p model at the departmental scale (Hauts-de-Seine).  The visualization relies on a dilation-erosion of classified buildings to reveal continuous zones associated with learned communities.  In light gray: majority community corresponding to the diffuse urban background.  In color: secondary communities highlighted by the model.  This figure aims to illustrate the spatial coherence induced by the topological constraint of the graph, without claiming a definitive morphological typology.}
    \label{fig:dmon3p}
\end{figure}

Figure~\ref{fig:dmon3p} is provided for illustrative purposes to provide a qualitative illustration of the structures induced by the learned assignments, without claiming exhaustive morphological interpretation at this stage.  Without introducing explicit geographic constraints, the learned communities form spatially continuous regions, indicating that the graph structure and topological weighting are sufficient to induce coherent local propagation of assignments. 

\FloatBarrier

\section{Conclusion}

This work proposes a methodological framework for unsupervised clustering of heterogeneous tripartite graphs, based on a differentiable formulation of modularity integrated into a DMoN-type end-to-end learning scheme. The main contribution lies in the definition of an optimizable tripartite modularity, efficiently evaluated through an exact factorization of co-paths, together with a structural flow normalization that enables the processing of graphs characterized by strong degree heterogeneity.

The ablation study demonstrates that this normalization is indispensable for optimization stability: in its absence, modularity maximization leads to degenerate regimes, such as collapse or artificial fragmentation of partitions. Large-scale experiments further show that the proposed framework is numerically stable, scalable to large graphs, and directly compatible with graph neural network architectures, without requiring prior projection or structural simplification.

Overall, these results confirm the relevance of the proposed approach for unsupervised learning on complex relational structures. This work opens several methodological perspectives, including extensions to inductive settings and the study of representation transferability across distinct graphs, which will be addressed in future work.

\bibliographystyle{unsrt}
\bibliography{modularite_tripartite}

\end{document}